\documentclass[a4paper,11pt]{article}

\usepackage{fourier}
\usepackage{color}
 \usepackage{graphicx}
\usepackage{url}
\usepackage[affil-it]{authblk}
\usepackage{amsmath}
\usepackage{wrapfig}
\usepackage{bbm, dsfont}
\usepackage{subcaption}
\usepackage{amssymb}
\usepackage{booktabs}
\usepackage{adjustbox}
\usepackage[T1]{fontenc}
\usepackage{times}

\usepackage[breaklinks,colorlinks]{hyperref}
\hypersetup{
    linkcolor=black,       
    citecolor=black,     
    filecolor=black,   
}
\begin{document}

\title{Vision-Language Model Based Handwriting Verification}

\author{%
    Mihir Chauhan\textsuperscript{††} \quad Abhishek Satbhai\textsuperscript{§} \quad Mohammad Abuzar Hashemi\textsuperscript{¶} \quad Mir Basheer Ali\textsuperscript{¶¶} \quad
    
    Bina Ramamurthy\textsuperscript{‡} \quad Mingchen Gao\textsuperscript{||} \quad Siwei Lyu\textsuperscript{†} \quad Sargur Srihari\textsuperscript{‡‡}\\ 
    
    Department of Computer Science and Engineering \\ 
    The State University of New York, Buffalo, NY, USA \\ 
    {\tt\small {\{mihirhem\textsuperscript{††}, bina\textsuperscript{‡}, mgao8\textsuperscript{||}, siweilyu\textsuperscript{†}, srihari\textsuperscript{‡‡}\}}@buffalo.edu} \\
    {\tt\small {\{ma.hashemi.786\textsuperscript{¶}, abhishek07satbhai\textsuperscript{§}, alimirbasheer\textsuperscript{¶¶}\}}@gmail.com}
}

\date{}
\maketitle
\thispagestyle{empty}

\begin{abstract}
Handwriting Verification is a critical in document forensics. Deep learning based approaches often face skepticism from forensic document examiners due to their lack of explainability and reliance on extensive training data and handcrafted features. This paper explores using Vision Language Models (VLMs), such as OpenAI’s GPT-4o and Google's PaliGemma, to address these challenges. By leveraging their Visual Question Answering capabilities and 0-shot Chain-of-Thought (CoT) reasoning, our goal is to provide clear, human-understandable explanations for model decisions. Our experiments on the CEDAR handwriting dataset demonstrate that VLMs offer enhanced interpretability, reduce the need for large training datasets, and adapt better to diverse handwriting styles. However, results show that the CNN-based ResNet-18 architecture outperforms the 0-shot CoT prompt engineering approach with GPT-4o (Accuracy: 70\%) and supervised fine-tuned PaliGemma (Accuracy: 71\%), achieving an accuracy of 84\% on the CEDAR AND dataset. These findings highlight the potential of VLMs in generating human-interpretable decisions while underscoring the need for further advancements to match the performance of specialized deep learning models. Our code is publicly available at: \url{https://github.com/Abhishek0057/vlm-hv}
\end{abstract}


\vspace{-0.6em}

\section{Introduction}
\label{sec:intro}
Handwriting verification is crucial in pattern recognition and biometrics, focusing on authenticating and identifying individuals by their handwriting. It is vital in forensics, where experts analyze samples to verify documents, identify forgeries, and provide court evidence.

\textbf{Historical Perspective:} Handwriting Verification methods \cite{shaikh2018hybrid} \cite{chauhan2019explanation} \cite{chauhan2024selfsupervisedlearningbasedhandwriting} have evolved from handcrafted features like GSC \cite{GSCfirst:5} to deep CNNs such as ResNet-18 \cite{Resnet:25}, and now Vision Transformers \cite{dosovitskiy2021image}, enabling comparison of both inter- and intra- writer variations in handwriting styles. 

\textbf{Challenges:} Despite the advancements, Forensic Document Examiners (FDE) remain skeptical due to the lack of interpretability in the model's decision making processes. Morever, these methods heavily depend on large datasets of labeled handwritten images, making dataset collection ($X(x_{q},x_{k}, y)$ with known $x_{k}$, questioned $x_{q}$ handwritten samples and corresponding writer labels $y$) is costly and time-intensive.

\textbf{Why use VLM?} Vision-Language Models (VLMs) integrate image and textual data to encode complex relationships between visual and linguistic information, enabling Forensic Document Examiners (FDEs) to interpret model decisions with clear, natural language explanations that enhance trust and reliability in forensic applications. They can adapt to the forensics domain in zero-shot scenarios without training examples and in few-shot scenarios with minimal examples by leveraging transfer learning capabilities, having been pre-trained on multiple tasks related to handwritten images and natural language understanding. Fine-tuning them with forensic-specific datasets further enhances their performance and applicability in real-world forensic use-cases. 

\textbf{VLMs applied to Forensics:} In \cite{scanlon2023ChatGPTForDigitalForensics}, the authors explore the capabilities of multimodal Large Language Models (LLMs) like GPT-4 in various digital forensics use-cases, including artifact understanding, evidence searching, anomaly detection, and education. More specific studies, such as \cite{jia2024chatgptdetectdeepfakesstudy}, demonstrate the effectiveness of GPT-4 specifically for the media forensics task of DeepFake image detection. Our research applies VLMs to handwriting verification. To our knowledge, we are the first to explore VLMs for the forensic task of handwriting comparison.



\begin{figure}[t]
  \centering
   \includegraphics[width=1\linewidth]{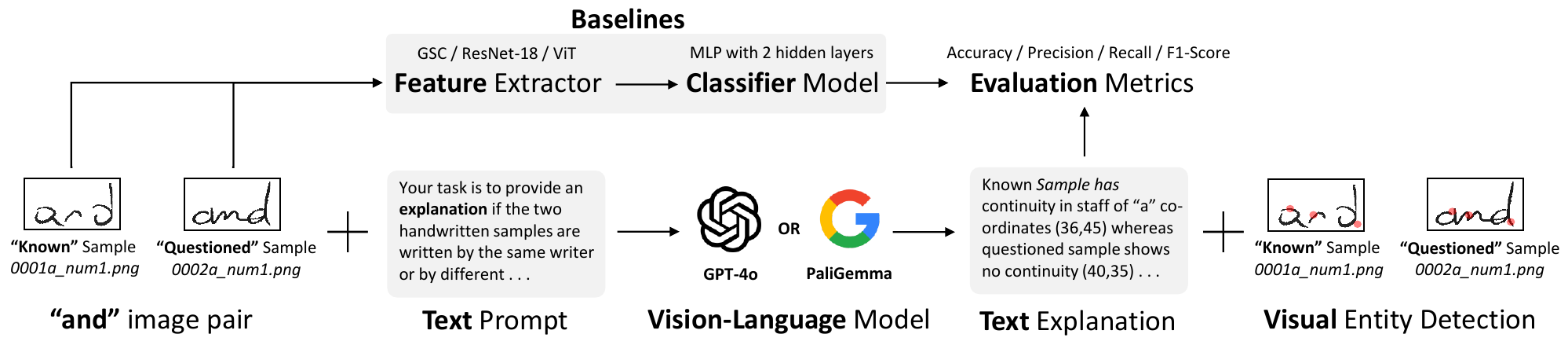}
   \caption{Overall process of evaluating Vision Language Model for Handwriting Verification against Baselines.}
   \label{fig:vlm_hv}
\end{figure}

\vspace{-0.8em}

\section{Methods}
Our study aims to use VLM in handwriting verification. We have chosen OpenAI’s GPT-4o VLM for its strong Visual Question Answering (VQA) capabilities. Using their API, we prompt GPT-4o \cite{gpt4o} with 0-shot Chain-of-Thought (CoT) reasoning. This approach allows us to first generate human-interpretable explanations and then determine whether the given questioned and known handwritten samples were written by the same person or by different writers, as shown in Figure \ref{fig:vlm_hv}. Since fine-tuning GPT-4o is not generally available, we compared its performance to a recent open-source VLM, PaliGemma \cite{beyer2024paligemmaversatile3bvlm}, using 0-shot prompt engineering and parameter-efficient supervised fine-tuning (PEFT) on a training dataset with 100 examples.  

\textbf{Data:} The experiments were conducted using 1,000 sample pairs of known and questioned images from CEDAR Letter \cite{Individuality:1} and CEDAR AND \cite{shaikh2018hybrid} dataset. CEDAR Letter dataset contains a letter manuscript written by 1568 writers three times. CEDAR AND dataset is a subset of CEDAR Letter data which only contains the lowercase handwritten word “\textbf{and}” extracted from full letter manuscript. The evaluation dataset includes 1,000 sample pairs of known and questioned images written by 368 writers with writer ids greater than 1200 (writer ids below 1200 were used to train the baseline models). Of these pairs, 500 are from the same writer and 500 are from different writers.

\vspace{-0.8em}

\section{Experiments \& Results}
\textbf{CEDAR AND Baselines:} As shown in Figure \ref{fig:vlm_hv} we use handcrafted features GSC features, CNN based ResNet-18 \cite{Resnet:25} and MaskedCausalVisionTransformer \cite{dosovitskiy2021image} (ViT) as our baseline feature extractors. GSC are 512-dimensional features extracted for the binarized ``AND'' images. All the three baselines were trained on 10\% and 100\% of known and questioned training pairs writers with writer ids less than 1200 resulting in 13,232 and 129,602 number of train pairs as shown in Table \ref{table:supervised_baseline}. The output of these feature extractors is fed into 2 fully-connected (FC) layers with 256 and 128 hidden neurons with ReLU activations. The final layer has 2 output neurons whose softmax activations represent similarity of samples with a one hot vector representation. We use categorical cross entropy loss given one-hot encoded logits compared to the target which is binary. 

\begin{table}[t!]
\centering
\begin{adjustbox}{width=1\textwidth}
\small
    \begin{tabular}{l|l|l|c|c|c|c|c}
    \toprule \toprule
    \textbf{Data} & \textbf{Model} & \textbf{Approach} & \textbf{\#Train Pairs} & \textbf{Accuracy} & \textbf{Precision} & \textbf{Recall} & \textbf{F1-Score} \\ \midrule
    CEDAR AND & GSC    & Supervised Training  & 13,232 & 0.71 & 0.69 & 0.72	& 0.69  \\
    CEDAR AND & ResNet-18 & Supervised Fine-Tuning & 13,232 &  0.72	& 0.70 & 0.73 & 0.72 \\
    CEDAR AND & ViT & Supervised Fine-Tuning & 13,232 & 0.65 & 0.68 & 0.64 & 0.66 \\ 
    CEDAR AND & GSC    & Supervised Training  & 129,602 & 0.78	& 0.81	& 0.77	& 0.79  \\
    \textbf{CEDAR AND} & \textbf{ResNet-18} & \textbf{Supervised Fine-Tuning} & \textbf{129,602} &  \textbf{0.84}	& \textbf{0.86}	& \textbf{0.82}	& \textbf{0.84}  \\
    CEDAR AND & ViT & Supervised Fine-Tuning & 129,602 & 0.79	& 0.80   & 0.78   & 0.79  \\
    \midrule
    CEDAR AND & GPT-4o      & 0-Shot CoT Prompt Engineering  & 0 & 0.7	&  0.68 & 0.7  & 0.69  \\
    CEDAR AND & PaliGemma   & 0-Shot CoT Prompt Engineering  & 0 & 0.65	&  0.66 & 0.65  & 0.65  \\
    CEDAR AND & PaliGemma   & Supervised Fine-Tuning  & 100 & 0.71	&  0.72 &  0.71 &  0.72 \\ \midrule
    \textbf{CEDAR Letter} & \textbf{GPT-4o}      & \textbf{0-Shot CoT Prompt Engineering}  & \textbf{0} & \textbf{0.65}	& \textbf{0.67}  & \textbf{0.65}  &  \textbf{0.66} \\
    CEDAR Letter & PaliGemma   & 0-Shot CoT Prompt Engineering  & 0 & 0.58	&  0.6 &  0.59 & 0.59  \\
    CEDAR Letter & PaliGemma   & Supervised Fine-Tuning  & 100 & 0.64 & 0.63  &  0.66  &  0.64 \\
    \bottomrule \bottomrule
    \end{tabular}
\end{adjustbox}
\caption{Performance metrics on 1000 sampled pairs of known and questioned pairs from evaluation dataset. Baseline performance for GSC, ResNet-18 and ViT is shown with 10\% \& 100\% of train writers on CEDAR AND dataset. 0-shot performance of GPT-4o and PaliGemma is evaluated for CEDAR AND and Letter dataset. Also, performance of fine-tuned PaliGemma on CEDAR AND dataset is observed. }
\label{table:supervised_baseline}
\end{table}

\textbf{Prompt Engineering with GPT-4o and PaliGemma:} To effectively utilize the GPT-4o VLM for handwriting verification, we experimented with various prompts to optimize both the generation of human-interpretable explanations and the accuracy of verification decisions. Initially, we crafted prompts that directed the model to compare specific features of the handwriting samples, such as stroke width, slant, and letter spacing. For example, prompts like "Describe the similarities and differences in the stroke patterns between the two samples" and "Identify any matching unique characteristics in the handwriting" were used. However, we observed significant variance in the model’s responses, leading to inconsistent results. To address this, we adopted a Chain-of-Thought (CoT) reasoning approach. This method guides the model through a structured reasoning process, allowing it to generate more consistent and reliable explanations. Since no examples were provided to VLM in the prompt engineering phase, the number of training pairs is observed as 0 in Table \ref{table:supervised_baseline}. 


Additionally, we utilized prompt engineered both VLMs to identify and mark coordinates of similarities and dissimilarities within the two images. By prompting the model to highlight specific areas of interest, such as matching or differing stroke patterns, we were able to generate detailed coordinates that pinpointed these features as shown in Figure \ref{fig:vlm_hv}. This approach provided clear visual indicators of the reasoning behind each verification decision. By incorporating CoT reasoning, presenting known and questioned samples as different images, and using VLM to identify key coordinates, we achieved a balance where the model could generate detailed, interpretable explanations while accurately determining the authenticity of the handwriting samples.

\textbf{Supervised Fine-Tuning with PaliGemma:} 
We fine-tuned the pre-trained PaliGemma model on 100 curated examples of 0-shot CoT prompt engineered PaliGemma results. The PaliGemmaProcessor within Transformers library was loaded to process the inputs which involved the preparation of prompt templates and batching text inputs with images. The image tokens and pad tokens were set to -100 to be ignored by the model, and these preprocessed inputs were used as labels for the model to learn from. The model, PaliGemmaForConditionalGeneration, was loaded and set to fine-tune only the text decoder, leaving the vision encoder and multimodal projector frozen. We used BitsAndBytes config options with 4bit quantization and LoRA \cite{hu2021loralowrankadaptationlarge} to train only 11M parameters out of the 3B parameters of the network. The results of supervised fine-tuned PaliGemma on both CEDAR AND and Letter datasets are shown in Table \ref{table:supervised_baseline}.

\textbf{Results:} The results of using VLMs for handwriting verification are presented in Table \ref{table:supervised_baseline} on the CEDAR AND and CEDAR Letter datasets. For the CEDAR AND dataset, CNN based ResNet-18 architecture outperforms 0-shot CoT prompt engineering approach with GPT-4o as well as supervised-finetuned PaliGemma with 100\% training pairs which suggest that although VLMs has a lot of potential for generating human interpretable decision making for the task of handwriting verification, still lags behind CNNs which aare fine-tuned on specific task of handwriting verification. This gap highlights the need for further improvements in fine-tuning regime of VLMs to enhance their effectiveness and reliability for specialized tasks like handwriting verification. For the CEDAR Letter dataset, the 0-shot CoT prompt engineering with GPT-4o also shows relatively higher performance than the supervised fine-tuning of PaliGemma but we observe high variability in the performance metrics because of low sample size. 

\section{Conclusion}
Our study explores the application of VLMs, specifically GPT-4o and PaliGemma, in the domain of handwriting verification. By leveraging the robust VQA capabilities of these models and employing 0-shot CoT reasoning through prompt engineering, we aimed to generate human-interpretable explanations for model decisions. Our experiments demonstrated that while VLMs offer significant improvements in interpretability and adaptability to diverse handwriting styles, they currently lag behind CNN-based architectures such as ResNet-18 in terms of performance. Specifically, ResNet-18 achieved an accuracy of 84\% on the CEDAR AND dataset, outperforming GPT-4o's 70\% accuracy and PaliGemma's 71\% accuracy. These findings suggest that while VLMs hold great promise for enhancing transparency and trustworthiness in forensic handwriting verification, there is still a need for further advancements in their fine-tuning regimes to improve their effectiveness and reliability for specialized tasks. Moving forward, we aim to work with forensic document examiners to create a fine-tuning dataset of explanation reports using text and visual information to ensure the applicability of our approach in practical forensic settings.



\bibliographystyle{apalike}
\bibliography{references}

\end{document}